# 2801 | In Situ Soil Property Estimation for Autonomous Earthmoving Using Physics-Infused Neural Networks

W. Jacob Wagner [a,b,*], Ahmet Soylemezoglu [b], Dustin Nottage [b], Katherine Driggs-Campbell [a]

[a] Department of Electrical and Computer Engineering, University of Illinois at Urbana-Champaign, Urbana, IL, United States

[b] Construction Engineering Research Laboratory, Engineer Research and Development Center, United States Army Corps of Engineers, Champaign, IL, United States

* Corresponding author: William.J.Wagner@erdc.dren.mil

## ABSTRACT

A novel, learning-based method for in situ estimation of soil properties using a physics-infused neural network (PINN) is presented. The network is trained to produce estimates of soil cohesion, angle of internal friction, soil-tool friction, soil failure angle, and residual depth of cut which are then passed through an earthmoving model based on the fundamental equation of earthmoving (FEE) to produce an estimated force. The network ingests a short history of kinematic observations along with past control commands and predicts interaction forces accurately with average error of less than 2kN, 13% of the measured force. To validate the approach, an earthmoving simulation of a bladed vehicle is developed using Vortex Studio, enabling comparison of the estimated parameters to pseudo-ground-truth values which is challenging in real-world experiments. The proposed approach is shown to enable accurate estimation of interaction forces and produces meaningful parameter estimates even when the model and the environmental physics deviate substantially.

## 1. Introduction

Autonomous earthmoving is important for enhancing the safety, productivity and efficiency of construction and mining operations, especially in hazardous and remote environments where human operators face significant risks. One big challenge in designing these systems is enabling the equipment to operate efficiently and robustly across a wide variety of soils. Soils are complex heterogeneous granular materials whose properties vary significantly, not just between disparate locations, but also locally. As soil strength properties vary, the force required to shear the soil changes as well, which has a large effect on the earthmoving process. For example, in low strength soils a bulldozer can fill its blade in a short distance by making a deep cut without the vehicle stalling or tracks slipping. In contrast, cut depth is limited for higher strength soils because force limits of the machine are exceeded by higher required shear forces that increase with depth.

### 1.1.1 Autonomous Earthmoving

Early approaches to autonomous excavation are posed as trajectory control problems where a desired trajectory is generated, typically based on some heuristic such as ensuring the swept volume is equal to the capacity of the bucket, and tracked using a position controller (Bradley and Seward 1998). However, these kinematic trajectory control approaches fail when the generated trajectory is infeasible due to force limits of the machine. The simplest approach to address this problem is to assume the soil is very high-strength and to conservatively select trajectories that ensure that the machine does not stall and tracks do not slip during execution of the motion. Unfortunately, this is very inefficient in lower-strength soils and limits the practicality of autonomous earthmoving.

To address these issues, alternative control strategies have been proposed that enable adaptation to changing soil parameters. In bulldozers, blade control technologies that adjust blade position to ensure the machine is exerting a force near the optimal load without producing track slip are commercially available (Hayashi et. al. 2013; Jackson 2017). Others have proposed impedance control (Ha et. al. 2000) or power maximization (Sotiropoulos and Asada 2019) as a way to balance the position control objective with force limitations. Instead of following a kinematic trajectory, a prototypical force-torque trajectory can be tracked which produces kinematic trajectories that vary depending on soil composition (Jud et. al. 2017). More recently, reinforcement learning methods have been used to train control policies that achieve greater fill-factors in excavation and scooping tasks (Azulay and Shapiro 2021; Backman et. al. 2021; Egli et. al. 2022).

However, local adaptation to soil conditions does not ensure efficient completion of a global terrain shaping task, implying that adaptation should be considered at the planning level as well. For an excavation task, iterative learning control has been used to adjust the desired trajectory between dig cycles to generate more feasible paths that result in increased bucket fill (Maeda and Rye 2012). Another approach is to estimate soil properties in situ and use these estimated properties to inform planning and control (Singh 1995a). At the planning level, these estimated properties can be used in combination with a model of earthmoving to ensure generated trajectories are feasible and efficient (Singh 1995b). To improve tracking of the trajectory, the interaction forces can be predicted and used with impedance control (Tan 2005).

*1.1.2  The Fundamental Equation of Earthmoving*

The first step in estimating soil properties is to assume a model, as the properties are tautologically tied to a soil model. In most soil property identification for autonomous earthmoving work, the Mohr-Coulomb model of soil shear strength is assumed which relates the shear strength of the soil, $\tau$, to the applied normal force, $\sigma_n$ via

$$\tau = c + \sigma_n \tan(\phi) \qquad (1)$$

where the parameters $c$ and $\phi$ are the soil cohesion and angle of internal friction respectively.

Autonomous earthmoving requires a model of how the soil properties affect this process. Using the method of trial wedges where soil is assumed to fail along a plane producing a wedge, (Mckyes 1989; Reece 1964), an equation can be derived that describes the applied force $F$ required to shear soil with a flat blade or bucket moving horizontally through the soil, as depicted in Fig. 1. The remaining forces acting on this soil wedge include the force of the loose soil accumulated on the blade which is referred to as surcharge or $Q$, the weight of the soil wedge $W$, the frictional component of the soil shear force combined with the normal force $R$, the soil cohesive force, and the soil-tool adhesive force. The forces are assumed to be in equilibrium, i.e. neither the soil nor the blade is accelerating, and are summed up along $\vec{x}$ and $\vec{z}$ directions to arrive at the fundamental equation of earthmoving (FEE):

$$F = f_{FEE}(\Theta) = \gamma d^2 w N_\gamma + cdw N_c + Q N_Q + c_a w N_a \qquad (2)$$

where the coefficients are given by

$$N_\gamma = \frac{[\cot(\rho)+\cot(\beta)]\sin(\alpha+\phi+\beta)}{2\sin(\eta)}, \quad N_c = \frac{\cos(\phi)}{\sin(\beta)\sin(\eta)} \qquad (3\text{-}6)$$
$$N_Q = \frac{\sin(\alpha+\phi+\beta)}{\sin(\eta)}, \quad N_a = \frac{-\cos(\rho+\phi+\beta)}{\sin(\rho)\sin(\eta)}$$

where $\eta = \delta + \rho + \phi + \beta$ is defined for notational convenience. This particular formulation of the FEE was developed by Holz et. al. (2013) to enable consideration of inclined surfaces. Initially, the equation has three unknown variables ($F, R, \beta$), but applying the equilibrium conditions reduces this down to the single unknown soil failure angle $\beta$. Typically, it is assumed that the optimal soil failure angle should be chosen such that

$$\beta^* = \arg\min_\beta N_\gamma \qquad (7)$$

In some cases, it is possible to derive a closed form expression for $\beta^*$, but often numerical methods are relied upon. The interaction force can be broken into its $\vec{x}\vec{z}$ subcomponents by

$$\mathbf{F} = f_{FEE}(\Theta) = \\ [F\cos(90° - \rho - \delta + \alpha),\; F\sin(90° - \rho - \delta + \alpha)] \qquad (8)$$

During bulldozing, the blade may move vertically in addition to horizontally as the desired cut depth is adjusted. This causes a change in the relative movement between the soil wedge and the blade. In the case that the wedge is moving down with respect to the blade, this results in the soil-tool adhesive and frictional forces changing direction. In order to account for these effects, Holz et. al. (2013) make the following modification

$$\delta' = \tanh(-C_1(\vec{\iota_b}\cdot\vec{v}))\delta, \; c_a' = \tanh(-C_1(\vec{\iota_b}\cdot\vec{v}))c_a' \qquad (9\text{-}10)$$

where $C_1$ is a positive constant, $\vec{\iota_b}$ is the unit vector pointing upwards along the blade, and $\vec{v}$ is the velocity vector of the blade represented in the $\vec{x}\vec{z}$ coordinate system. The values $\delta'$ and $c_a'$ are used in place of $\delta$ and $c$ in equations 2-6.

*1.2  Soil Property Estimation*

There has been considerable attention on developing methods for estimating and predicting vehicle traversability for the purposes of autonomous off-road navigation (Borges et. al. 2022). Of particular interest for this work is the characterization of soil properties from on-board vehicle sensing including vision and proprioception, sometimes referred to as inverse terramechanics (IT) estimation. In-situ soil property estimation using these methods has been studied within the field of terramechanics and is often motivated by the need for autonomous navigation of planetary rovers (Lopez-Arreguin et. al. 2021). This problem is formulated as the estimation of parameters of a soil-wheel interaction model and is often solved using optimization-based approaches (Iagnemma et. al. 2004), although filtering methods have been developed (Dallas et. al. 2020) and learning based methods have been used (Lopez-Arreguin and Montenegro 2021).

For autonomous earthmoving tasks, a similar IT approach can be taken. However, as the vehicle is more forcefully interacting with the terrain, different models must be assumed. Typically, a subset of the model parameters is assumed to be known, and the remaining parameters are found by fitting the model to observed interaction forces (Singh 1995a, Luengo et. al. 1998, Tan et. al. 2005, Althoefer et. al. 2009). It is often difficult to determine the accuracy of the estimated parameters directly, but Tan et. al. (2005) report predictions of the soil failure force to within 10%-30%.

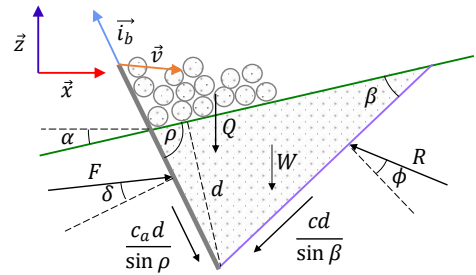

**Fig. 1.** FEE soil wedge geometry and force diagram. This is a cross-sectional view depicting a blade, shown in gray, moving in the $\vec{v}$ direction through the soil at a depth of cut $d$ and failing the soil along the purple line at an angle $\beta$ with respect to the surface. The soil surface, shown in green, is inclined at the angle $\alpha$ from the horizon and the flat blade is at an angle $\rho$ with respect to the soil surface. The passive failure case is assumed where the soil wedge moves up with respect to the blade and forward with respect to the un-sheared soil. The cohesive forces resist this movement and are drawn accordingly. (Holz et. al. 2013)

*1.3  Contributions*

The estimation of soil properties is a key challenge for autonomous earthmoving operations, such as bulldozing. Existing IT methods have mainly focused on the soil-tire interaction for vehicle navigation applications and, while motivating, these methods do not directly apply to earthmoving tasks due to different underlying physics. The relatively limited prior work in IT methods for construction tasks leverage optimization approaches to find parameters of the earthmoving model given observations of the interaction force. In this work, we propose a novel IT method based on physics-infused neural networks (PINNs) that can estimate soil properties in situ from kinematic and control observations of a bulldozer blade. The main contributions of this work are:

- Development of a PINN-based in situ soil-property estimation method that incorporates physical laws and constraints into the learning process without requiring knowledge of true soil properties
- Evaluation of the accuracy and robustness of the proposed method using simulation data with ground-truth soil properties
- Introduction of a physical inconsistency loss function that enables enforcement of minimization or maximization of certain physical quantities
- Demonstration of a novel technique to estimate the blade-soil interaction forces directly from kinematic and control data

## 2. Methods

The creation of a simulation to study the in-situ soil-property estimation problem is first discussed, and the PINN approach to soil property estimation is then developed.

### 2.1 Simulation

In this work, the proposed soil property estimation method is evaluated to determine how capable the system is at extracting characteristics of the soil. Real world evaluation a challenging task, as evaluation of the full system would require significant instrumentation of a bulldozer, time consuming real-world experimentation across various soil types and levels of compaction, and careful setup (e.g., compaction, moisture content) and characterization of these soils to establish a soil property ground-truth to enable validation of the approach. Instead, using a simulation as an analog for experiments on a real bulldozer is proposed. To perform these experiments, a simulator that exhibits similar behavior to a real system is required, but faithful representation of the dynamics of a specific piece of earthmoving equipment for a specific soil is not strictly necessary.

### 2.1.1 Vortex Studio

Vortex Studio[1] is a simulation software which enables modelling of multi-body dynamics and supports accurate real-time soil-tool interaction physics. In Vortex, the soil is modeled using a hybrid heightfield-particle representation. An FEE based model is used to determine the reaction force of the soil prior to shearing. As the relative density $I_d$ of the soil increases, $\phi$ and $c$ are varied to increase the resistance of the soil to shearing due to stronger particle interlocking, and $\gamma$ is varied to adjust the weight of the soil wedge and surcharge (Holz 2009).

When the force exerted by the tool on the soil reaches this threshold, the portion of the heightfield in contact with the blade is converted into particles whose behavior is governed by a discrete element model (DEM) simulation which more accurately characterizes the behavior of the disturbed soil. As particles come to rest, they are then reintegrated back into the heightfield representation. The initial relative density is defined as a fixed value but may vary throughout the simulation as the soil is compacted by the tool or as sheared particles are merged back into the heightfield (Holz 2009, Holz et. al. 2013). The remainder of the soil properties remain constant for a given soil type and do not vary within a simulation episode.

Haeri et al. (2020) showed that Vortex Studio is capable of predicting the soil-tool interaction forces for a lunar simulant material within 20%-30% of the measured force obtained from real-world experimentation. They introduce some modifications to the original Vortex model, including a term that reduces the contribution of the surcharge by a factor of 10 in equation 2. Additionally, they find that selecting $\beta$ such that the top surface of the soil wedge matches extent of the surcharge pile, better matches the experimental data as compared to minimization of $N_\gamma$ as in Equation 7.

One benefit of using a simulation to study soil-property estimation is that it can provide ground-truth knowledge of the soil parameters for validation of a given approach. However, in Vortex, this is not straightforward as the simulation is not a purely physics-based model, but rather a combination of physics and heuristics that have been added to improve the realism, stability, and performance of the system. In addition to the modifications discussed by Haeri et. al. (2020), these heuristics include various adjustments to the FEE force calculation, such as limiting the force based on the submerged surface area, the addition of penetration forces, scaling of the FEE force, inclusion of additional frictional, elastic, and damping contact forces between the heightmap and tool, etc. These heuristics are not directly derived from the soil properties, but rather must be tuned to match experimental data or may be modified to incorporate operator feedback, improving the realism of the simulation.

Vortex provides four default soil configurations (Clay, Loam, Sand, and Gravel), but further tuning is recommended for specific applications. However, it is not obvious how to configure these parameters to achieve realistic performance for different soil types and blade geometries. Despite these limitations, Vortex meets the needs for a system that exhibits realistic earthmoving phenomenon and is sufficiently complex to act as an analog for experiments on a real machine.

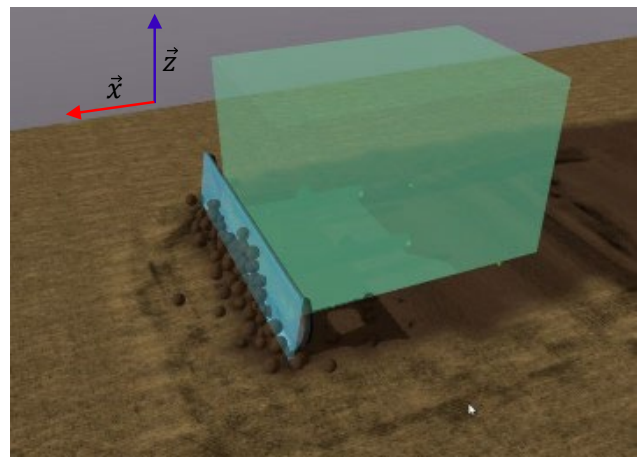

**Fig. 2.** Rendering of simple bladed vehicle simulation. The chassis is shown in green and the flat blade collision geometry is shown in blue. The soil particles generated due to the blade failing the soil constitute the surcharge.

### 2.1.2 Simple Bladed Vehicle Model

A simplified simulation of a bladed earthmoving vehicle is developed, see Figure 2, and consists of two dynamic components: a chassis and a blade. The chassis is modeled as a rectangular prism (3m, 2m, 2m) with a mass of 5,000kg. The blade is flat 3.164m wide, 0.660m tall, 0.001m thick[2] with a mass of 400kg.

---

[1] https://www.cm-labs.com/en/vortex-studio/

Vortex constraints, which enable restricting the motion between two or more parts, are used to control the motion of the chassis and the blade. Motorized constraints enable control of a part's velocity and exert a force proportional to the velocity error. The inverse of the loss parameter specifies this constant. Locked constraints enable control of a part's position and exert a viscoelastic force on the part equivalent to a spring-damper system (CM Labs Simulations, 2016). A motorized constraint is used to control the forward velocity of the chassis and a locked constraint is used control its vertical position. The lateral position and orientation of the chassis are locked, limiting the motion of the vehicle to the $\vec{x}\vec{z}$ plane. The blade is attached to the chassis body using a locked constraint enabling control of the vertical position relative to the chassis.

To emulate limits on a real machine, the relative position of the blade with respect to the chassis is limited to enable the bottom edge of the blade to reach 0.3m. below and 1.0m above the surface when the chassis is on level ground. The soil height is measured at four locations underneath the chassis and the average of these height is used to command the chassis vertical position to emulate a simple vehicle suspension. Additionally, the vertical and horizontal constraint forces are limited to 30kN and 20kN respectively to emulate the tractive and penetrative force limits of a bladed earthmoving machine.

On a bulldozer, hydraulic pressures in the vehicle hydrostatic drive circuit and in the cylinders controlling the blade position can be measured to derive estimates of the tractive force given knowledge of the machine's dynamics (Yamamoto et. al. 1997). The simulation developed in this effort has a simpler drive system, therefore, the simulator chassis-blade constraint forces are used to measure the interaction force instead. Significant effort was given to tune the loss, stiffness, and damping coefficients of the constraints to yield a stable simulation with minimal oscillations in the constraint forces and part positions. However, oscillations on the order of 5kN and 10kN in the x and z forces respectively are commonly observed in the experiments, see Fig. 3. These oscillations arise from the interaction of the constraints used to control the vehicle and the constraints imposed by the soil-tool interaction model. In particular, when a portion of the heightfield is converted into particles, the blade momentarily loses contact with the heightfield resulting in a significant disturbance in the force applied to the blade. This amounts to a step input to the control constraints and can lead to transient violations of the force limits, as Vortex constraints are essentially dynamical systems.

During simulation, at a rate of 60Hz, observations are collected of the blade $x$ and $z$ positions and chassis $z$ positions,

---

[2] Originally, the blade was made to have a more reasonable thickness, but initial experimentation revealed problems with the vehicle being able to penetrate the soil surface even with large forces for denser soil configurations. The authors believe that this is an artifact of the way contact detection is used to compute the soil blade angle ρ internal to Vortex. The hypothesis is that the system computes the angle using the bottom surface of the blade instead of the leading edge causing the simulator to produce unrealistically large reaction forces. Reducing the blade width alleviated most of these problems.

$\boldsymbol{p} = [p_x^b, p_z^b, p_z^c]$, the blade $z$ velocity and chassis $x$ and $z$ velocities, $\boldsymbol{v} = [v_z^b, v_x^c, v_z^c]$, and the $x$ and $z$ cutting forces, $\boldsymbol{F} = [f_x^b, f_z^b]$. Additionally, the commanded blade relative and absolute $z$ position, and the chassis $x$ velocity $\boldsymbol{u} = [u_{z_r}^b, u_{z_a}^b, u_x^c]$ are collected. At each timestep $t$, the observations and actions are concatenated into the observation vector $\boldsymbol{o_t} = [\boldsymbol{p_t}, \boldsymbol{v_t}, \boldsymbol{u_t}]$ which is consumed by the PINN. Pseudo-ground truth soil-properties are collected from the simulation as $\boldsymbol{\Theta} = [\phi, c, \delta, c_a, \gamma, \rho, \alpha, w, d, Q, \vec{v}]$, where $Q$ is obtained using the Vortex soil mass sensor plugin. The simulation does not provide access to the soil failure angle $\beta$, but for notational simplicity it will be included in $\boldsymbol{\Theta}$ for the remainder of the text.

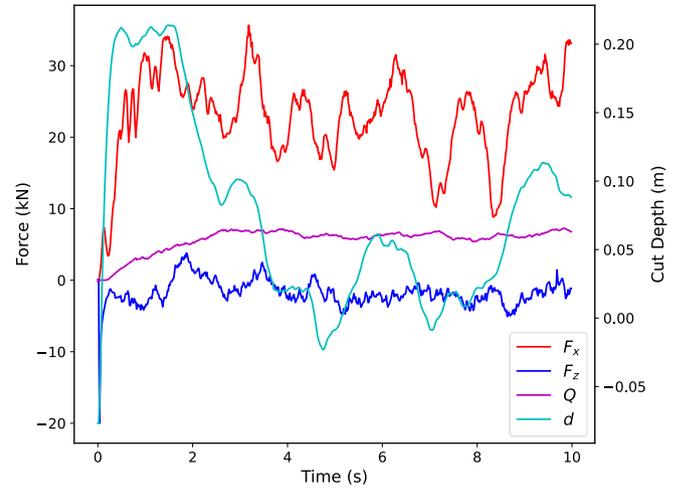

**Fig 3.** Measured forces and cut depth for a typical episode within the $D_{\text{default}}$ dataset. This episode was recorded for a loam soil with $I_d = 60\%$ and illustrates the large amplitude noise in force measurements observed particularly for denser soils. Also, note how the surcharge grows with time indicating the accumulation of soil on the blade from the cutting operation.

### 2.1.3 Data Collection Controller

To ensure clean data is collected to train an effective model, a simple data collection controller was designed. In the case that the force required to shear the soil is larger than the machine limits, the blade becomes stalled, and the soil is not sheared. Therefore, an anti-stall blade controller is developed which enables automatic data collection.

For a given desired forward velocity $\bar{v}_x$ and target absolute cut depth $\bar{d}_z$, the controller adjusts the blade relative position command $u_{z_r}$ to ensure that the velocity tracking error $e_{v_x}$ does not drop below a target velocity dependent threshold $e_{v_{min}} = \bar{v}_x c_{v_x}$. This is accomplished using a gain-scheduled integral controller which reduces the desired depth of cut by a depth offset $\Delta \bar{d}_z^k$ at each timestep $k$; this quantity is limited to enable raising of the blade only a small amount above the surface. The controller is defined as follows

$$\Delta \bar{d}_z^k = K_{\Delta p}(e_{v_x}^k - e_{v_{min}}) + \Delta \bar{d}_z^{k-1} \quad \in [0, \bar{d}_z + \Delta \bar{d}_{z_{max}}] \quad (11)$$
$$u_{z_r} = -(\bar{d}_z - \Delta \bar{d}_z^k) \quad (12)$$

The scheduled depth offset gain term, $K_{\Delta p} = K_v/\bar{v}_x$, modulates the depth offset integral gain $K_v$ to enable improved forward velocity tracking for low desired velocities $\bar{v}_x$. Overall, the effect of this controller is that vehicle maintains forward motion with some low-frequency oscillation in the depth of cut and forward velocity around the targets. As the soil becomes harder to shear and vehicle force limits are reached, the desired depth of cut becomes harder to track, but stalling is prevented ensuring the soil failure condition assumed by the FEE is met.

## 2.2 Physics-Infused Neural Network (PINN)

Broadly, the approach taken in this effort is to train a neural network to produce estimates of the soil parameters which are then fed through the FEE model to produce an estimated force. The network ingests a $T$ step history of observations and control commands $\boldsymbol{o}_{[t:t+T]}$ along with known FEE parameters $\boldsymbol{\Theta}_k = [c_a, \gamma, \rho, \alpha, w, d, Q, \vec{v}]$, and outputs an estimate of the unknown parameters $\widehat{\boldsymbol{\Theta}}_u = [\hat{\phi}, \hat{c}, \hat{\delta}, \hat{\beta}, \Delta d]$ and the estimated interaction force $\widehat{\boldsymbol{F}}$. It is important to include control actions in addition to kinematic information, as tracking error can be thought of as a measure of the interaction force. For example, Bradley and Seward (1998) use the position tracking error to achieve "software force feedback" that can be used to determine when the bucket comes into contact with the ground.

By properly constraining the network, the hypothesis is that the model will converge to estimating reasonable soil parameters because this representation is the most parsimonious representation of the data. In other words, the model should learn to estimate meaningful soil properties as they explain the observations in the simplest form.

### 2.2.1 Network Architecture

While it may be possible to estimate FEE parameters from observations and commands at a single time-step, this is difficult given the noise present. Therefore, observations and actions across $T = 60$ timesteps (1 sec.) are incorporated, and a temporal convolution is leveraged to compress the $T \times 9$ dimensional observation $\boldsymbol{o}_{[t:t+T]}$ into a $l \times N$ vector, where $l$ and $N$ are model hyperparameters. This vector is passed through a stacked transformer style encoder network with multi-headed attention, modeled after the network presented by Zerveas et. al. (2021), producing another $l \times N$ vector. This vector is then combined with the known FEE parameters $\boldsymbol{\Theta}_k$ and passed through a dense network with 2 hidden layers of 20 neurons each that is referred to as the integration network. The role of this network is to incorporate the knowledge of the existing soil parameters with the latent state produced by the transformer from the observation history. This enables estimation of the unknown soil parameters. Without this network, the system cannot learn to account for variation in $\boldsymbol{\Theta}_k$ by changing $\widehat{\boldsymbol{\Theta}}_u$. A linear output layer produces the estimated unknown FEE parameters $\widehat{\boldsymbol{\Theta}}_u$ and the residual end to end force $\boldsymbol{F}_r$. The FEE parameters are combined $\widehat{\boldsymbol{\Theta}} = [\boldsymbol{\Theta}_k, \widehat{\boldsymbol{\Theta}}_u]$, clipped to remain within the limits defined in Table 1, and passed to the FEE network to produce the FEE reaction force $\widehat{\boldsymbol{F}}^{FEE} = f_{FEE}(\widehat{\boldsymbol{\Theta}})$. The residual and FEE reaction forces are then summed to produce the estimated interaction force $\widehat{\boldsymbol{F}} = \boldsymbol{F}^{FEE} + \boldsymbol{F}^r$. See Fig. 4 for a visual depiction of this architecture.

The residual depth of cut which augments the depth of cut by $d' = d + \Delta d$, was added to the model after initial experimentation revealed significant sensitivity of the FEE model to the depth of cut; major improvements in reconstructing the observed force entirely from the ground-truth parameters was shown to be possible with hand tuning of $\Delta d$. Similarly, the residual interaction force $\boldsymbol{F}^r$ was added to the network to avoid non-FEE components of the simulation (e.g. noise in the measured force and non-zero measured force when $d < 0$) corrupting the parameter estimates.

### 2.2.2 Network Training

To allow for stable training, observations $\boldsymbol{o}$ are normalized to have unit mean and variance where the normalizing mean and variances are computed from the training set observations. Values for the residual force are normalized $\boldsymbol{F}^r$ to 10% of the value of $\boldsymbol{F}$. Both known and unknown FEE parameters are scaled using min-max normalization with ranges provided in Table 1. The ranges for $\phi, c, \delta,$ and $c_a$ are obtained from multiple sources listing geotechnical parameters (Geotechdata.info 2023, MATHalino 2023, and Fine Software 2023) while the remaining parameters are scaled using knowledge of machine limits and geometry of the FEE. During tuning of the model, some liberty was taken to adjust parameter ranges for $\beta$ and $c$ from the initially assumed values in order to avoid the model converging to a poor local optimum.

**Table 1**
FEE parameter normalization ranges and limits.

| Parameter | Norm. Range | Limits $[l_l, l_u]$ | Units |
|---|---|---|---|
| $\phi$ | [17, 45] | [0, 90] | ° |
| $c$ | [0, 10e3] | [0, -] | Pa |
| $\delta$ | [11, 35] | [0, 90] | deg. |
| $c_a$ | [0, 10e3] | [0, -] | Pa |
| $\gamma$ | [14e3, 22e3] | [0, -] | N/m³ |
| $\rho$ | [2, 178] | [2, 178] | ° |
| $\alpha$ | [10, 10] | [-30, 30] | ° |
| $w$ | [0, 3.164] | [0, 3.164] | m |
| $d'$ | [0, 0.3] | [0, 0.660] | m |
| $\Delta d$ | [-5e-2, 5e-2] | [-5e-2, 5e-2] | m |
| $Q$ | [0, 10e3] | [0, -] | N |
| $v_x$ | [0,1] | [-2, 2] | m/s |
| $v_z$ | [-1, 1] | [-2, 2] | m/s |
| $\beta$ | [11.5, 34.5] | [11.5, 34.5] | ° |
| $\partial N_\gamma/\partial \beta$ | [-10, 10] | [-, -] | unitless |
| $\eta$ | [2, 178] | [2, 178] | ° |
| $\zeta = \phi - \delta$ | [11, 35] | [0, -] | ° |

The loss function for the network is comprised of five types of losses: a mean average error force prediction loss ($\mathcal{L}_{MAE}^{F}$), a MAE regularization of the residual force ($\mathcal{L}_{MAE}^{F_r}$), ReLU-based regularization of FEE parameters ($\mathcal{L}_{ReLU}^{\chi}$), a mean square error (MSE) based regularization of the residual depth of cut ($\mathcal{L}_{MSE}^{\Delta d}$), and a MAE regularization of the gradient $\partial N_\gamma/\partial \beta$ ($\mathcal{L}_{MAE}^{\partial N_\gamma/\partial \beta}$). It is defined as follows

$$\mathcal{L}(\boldsymbol{F}, \widehat{\boldsymbol{\Theta}}, \widehat{\boldsymbol{F}}, \boldsymbol{F}^r) = \lambda_F \mathcal{L}_{MAE}^{F}(\boldsymbol{F}, \widehat{\boldsymbol{F}}, \mathbf{w}_{xz}) + \lambda_{F_r} \mathcal{L}_{MAE}^{F_r}(\boldsymbol{F}^r, \mathbf{0}, \mathbf{w}_{xz}) + \sum_{\chi \in X} [\lambda_\chi \mathcal{L}_{ReLU}^{\chi}(\chi, \boldsymbol{l}^\chi)] + \lambda_{\Delta d_{MSE}} \mathcal{L}_{MSE}^{\Delta d}(\Delta \bar{d}, \Delta d) + \lambda_{\partial N_\gamma/\partial \beta} \mathcal{L}_{MAE}^{\partial N_\gamma/\partial \beta}\left(\frac{\partial N_\gamma}{\partial \beta}, \mathbf{0}, \mathbf{w}_{\partial N_\gamma/\partial \beta}\right) \quad (13)$$

$$\mathcal{L}_{MAE}^{y}(\bar{\boldsymbol{y}}, \boldsymbol{y}, \mathbf{w}) = \sum_{i=0}^{|y|} \mathbf{w}_i |\bar{\boldsymbol{y}}_i - \boldsymbol{y}_i| \quad (14)$$

$$\mathcal{L}_{MSE}^{\Delta d}(\Delta \bar{d}, \Delta d) = (\Delta \bar{d} - \Delta d)^2 \quad (15)$$

$$\mathcal{L}_{ReLU}^{\chi}(\chi, \boldsymbol{l}^\chi) = max(0, l_l^\chi - \chi) + max(0, \chi - l_u^\chi) \quad (16)$$

The FEE parameters regularized to remain within the limits listed in Table 1 are $X = [\phi, c, \delta, \beta, \eta, \zeta, \Delta d, d']$ and the weights $\lambda_\chi = 1$ for all of these parameters except for $\lambda_\eta, \lambda_\beta = 10$. These values are chosen to be higher because when violated, they lead to singularities in Eq. 2 causing $\widehat{\boldsymbol{F}}$ to become unstable, derailing training of the model. The other loss weights are tuned to achieve low force estimation error and low residual contributions. Their values are: $\lambda_F = 0.5, \lambda_{F_r} = 3e-2, \lambda_{\Delta d_{MSE}} = 5e-3, \lambda_{\partial N_\gamma/\partial \beta} = 0.2$.

Although not explicitly expressed in Eq. 13, averaging is performed over all samples in the batch for $\mathcal{L}^{\chi}_{\text{ReLU}}$, $\mathcal{L}^{\partial N_\gamma/\partial\beta}_{MAE}$, and $\mathcal{L}^{\Delta\delta}_{MSE}$. The $\mathcal{L}^{F}_{MAE}$ loss is only averaged over samples for which the limit constraints for $\eta$ and $\beta$ are not violated to avoid the gradient step causing large changes to the network parameters when near a singularity of the FEE model. Similarly, $\mathcal{L}^{F_r}_{MAE}$ is only averaged over samples for which the residual augmented depth of cut $d' > 0$ to avoid penalizing the residual force from compensating for when the FEE model produces a zero-output force. All of the losses are computed using the normalized values, which helps to expedite hyperparameter tuning.

To encourage the model to estimate $\beta$ so that it is consistent with the assumptions of the FEE model, it is limited to lie within an interval for which the minimum is expected to exist using ReLU-based physical inconsistency regularization, Equation 16. Additionally, $\partial N_\gamma/\partial\beta$ is regularized to be small. It is important to note that only gradients of this loss $\mathcal{L}^{\partial N_\gamma/\partial\beta}_{MAE}$ with respect to $\beta$ are allowed to flow, i.e. gradients with respect to the other model parameters of $\partial N_\gamma/\partial\beta$ are zero. Combining these losses encourages the network to converge to the critical point $\beta^*$ that is the global optimum of Eq. 7 in a *soft* fashion, allowing for deviations from this optimum when significant force prediction accuracy improvements can be achieved. This approach is novel and expands the type of physical inconsistency loss functions that can be accounted for when developing a PINN beyond equality and inequality constraints, as outlined by Karpatne et. al. (2022).

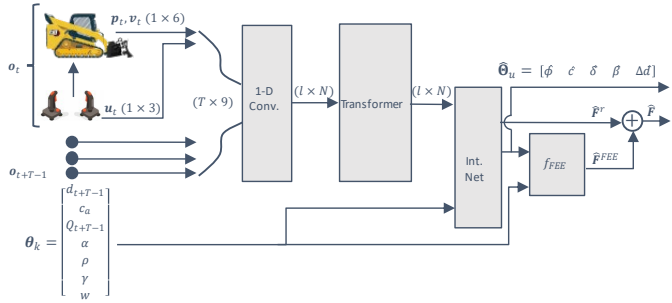

**Fig. 4.** FEE parameter estimation PINN diagram.

## 3. Results & Discussion

To evaluate the approach proposed in this effort, the PINN derived parameters are compared with the ground truth parameters used in the simulation.

### 3.1 Experimental Setup

Two sets of experiments using two different datasets collected from different configurations of the Vortex soil model are performed. The first dataset, $D_{\text{FEE}}$, is collected by configuring the soil model where the heuristics discussed in Section 2.1.1 are disabled such that only the reaction forces generated by the FEE model are exerted on the blade. Additionally, particle generation is disabled, i.e. $Q = 0$, to ensure that the forces generated by particles do not confound the parameter estimation. The goal of making these changes is to collect a dataset for which accurate ground-truth knowledge of the soil parameters is available. This enables evaluation of the accuracy of the parameter estimation approach in an ideal scenario, where the physics that governs the system matches the model that is assumed.

The second dataset, $D_{\text{default}}$, uses the default configuration for the soils, leaving the generation of particles and all of the heuristics enabled. In this case, there exists a significant mismatch between the physics governing the system and the assumed model. As discussed previously, while the simulation does not necessarily reflect the real-world physics to some quantifiable accuracy, the system exhibits similar phenomena as observed in real-world earthmoving operations. The $D_{\text{default}}$ dataset can therefore be treated as an analog to experimentation on a real-world system.

For all experiments, the ground is assumed to be flat and level, meaning that $\alpha = 0$ and $d = p_z^b$. The mass of the particles in front of the blade is obtained using a soil mass sensor yielding the surcharge force $Q$. A simple windowed average is used to filter the velocity measurement $\vec{v}_t = \text{avg}([v_x^c, v_z^b]_{t-10:t})$ to reduce oscillations in the velocity-based scaling of the soil-tool parameters. The blade angle is fixed for all experiments at $\rho = 80°$ as is the tool width at $w = 3.164$m. The initial values of $\phi, c,$ and $\gamma$ are obtained at the beginning of an episode based on the initial relative density $I_d$, as it is not supported to obtain the changing parameter values while running the simulation. This is not problematic in practice as the system only moves forward and does not interact with any previously disturbed terrain. The adhesion is fixed for all soil types, $c_a$=200 Pa, as is the soil-tool friction angle, $\delta = 10°$. Since the blade angle is fixed at $\rho = 80°$, the FEE predicts zero force in the $z$ direction for larger forward velocities. To enable observation of some non-zero forces in the $z$ direction for $D_{\text{FEE}}$, the default soil-tool angle is overridden to $\delta = 15°$. All of these values are collected to form the pseudo-ground-truth FEE parameters $\boldsymbol{\Theta}$.

The $D_{\text{FEE}}$ dataset consists of 4,320 sequences collected from 432 separate 10-second-long episodes using the default simulation update frequency of 60 Hz. Data is collected across all 4 default soil types and the relative density of the soil is varied across episodes, $I_d \in [0,100]$, as are target velocity and cut depth commands, $\bar{v}_x \in [0.3,1.0]$m/s, $\bar{d}_z \in [0.05,0.3]$m. The $D_{\text{default}}$ dataset contains 5,760 sequences from 576 episodes and is collected similarly. The PINN is implemented in PyTorch, training performed on a laptop equipped with a GPU, and the network loss in Eq. 13 is minimized using the Adam optimizer.

### 3.2 Results

The model does well predicting the interaction force, achieving an average magnitude error $|\boldsymbol{F} - \widehat{\boldsymbol{F}}|$ of approximately 730N, or 11% of the measured force, on $D_{\text{FEE}}$ and 1940N, 13% of measured, on $D_{\text{default}}$. To put this in perspective, normalizing the error by machine limits reveals 2% error on $D_{\text{FEE}}$ and 5% error on $D_{\text{default}}$. This is compelling and shows that the network is doing a good job of learning the dynamics of both the simplified and complex system. To evaluate the ability of the model to identify soil parameters, parameter estimates $\widehat{\boldsymbol{\Theta}}$ and estimated forces $\widehat{\boldsymbol{F}}^{FEE}$ are compared to their pseudo-ground-truth counterparts as a function of relative density. Parameter estimates from each episode within the dataset where relative density $I_d$ and soil type are the same are aggregated.

$$\overline{\boldsymbol{\Theta}}_i^j = [\widehat{\boldsymbol{\Theta}}^j \in \boldsymbol{D} \text{ s.t. } I_d = i] \text{ for } j \in \widehat{\boldsymbol{\Theta}} \qquad (17)$$

The mean and variance of the aggregated parameters are then computed at each $I_d$ for each soil type and visually depicted in the upper half of Figs. 5 and 6. This enables direct comparison between individual parameter estimates and their corresponding pseudo-ground-truth.

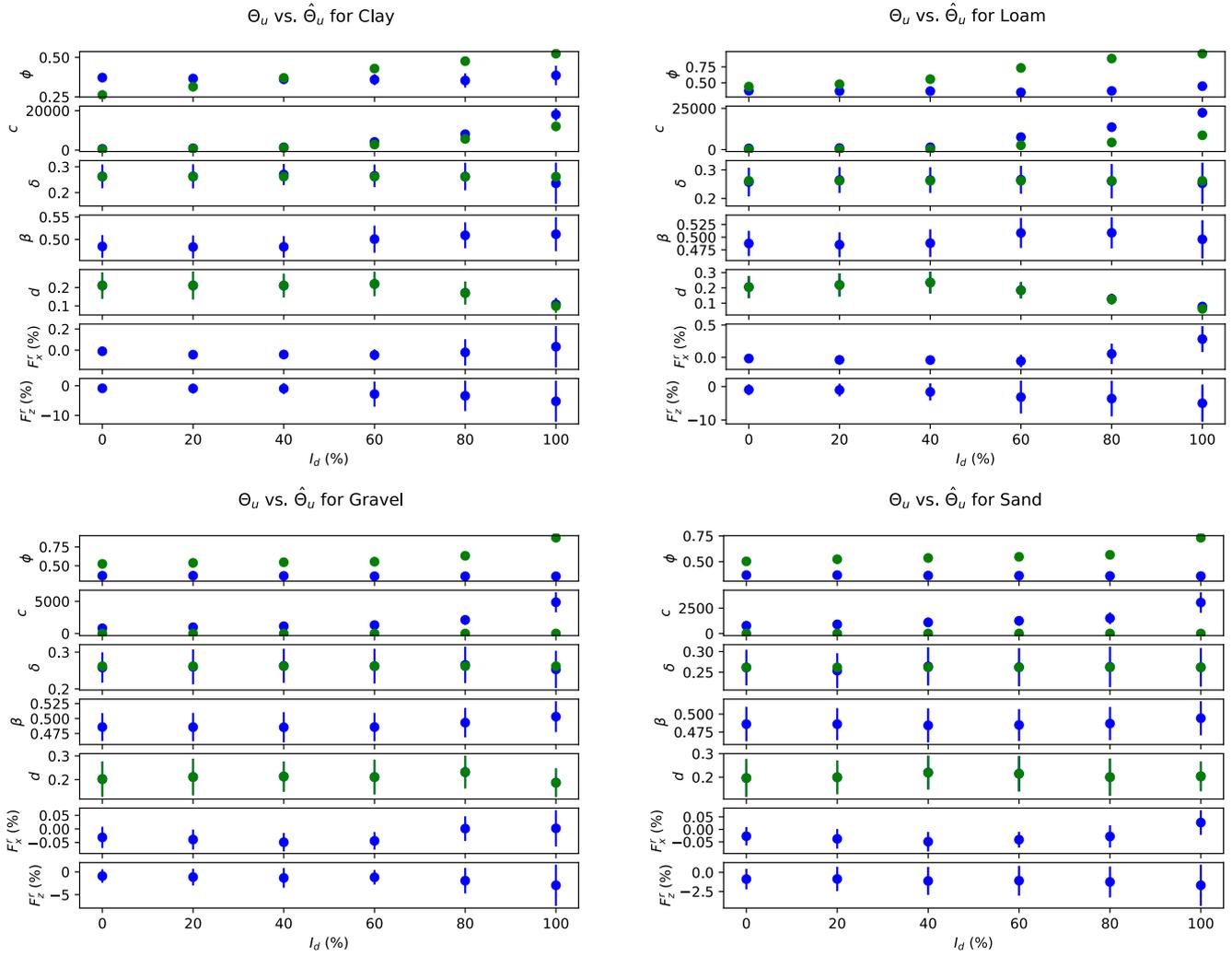

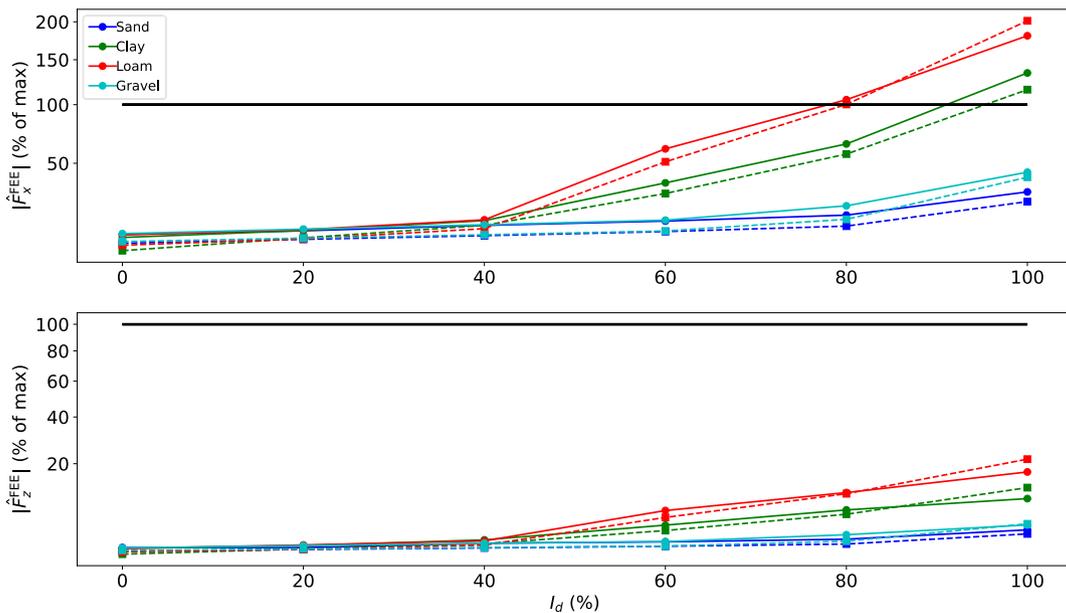

**Fig.. 5** $D_{FEE}$ dataset (top) FEE parameters estimated by the model (blue) and the pseudo-ground-truth values obtained from Vortex. (green), (bottom) Soil failure force predicted by the FEE component of the model. Units for all angles are in radians,

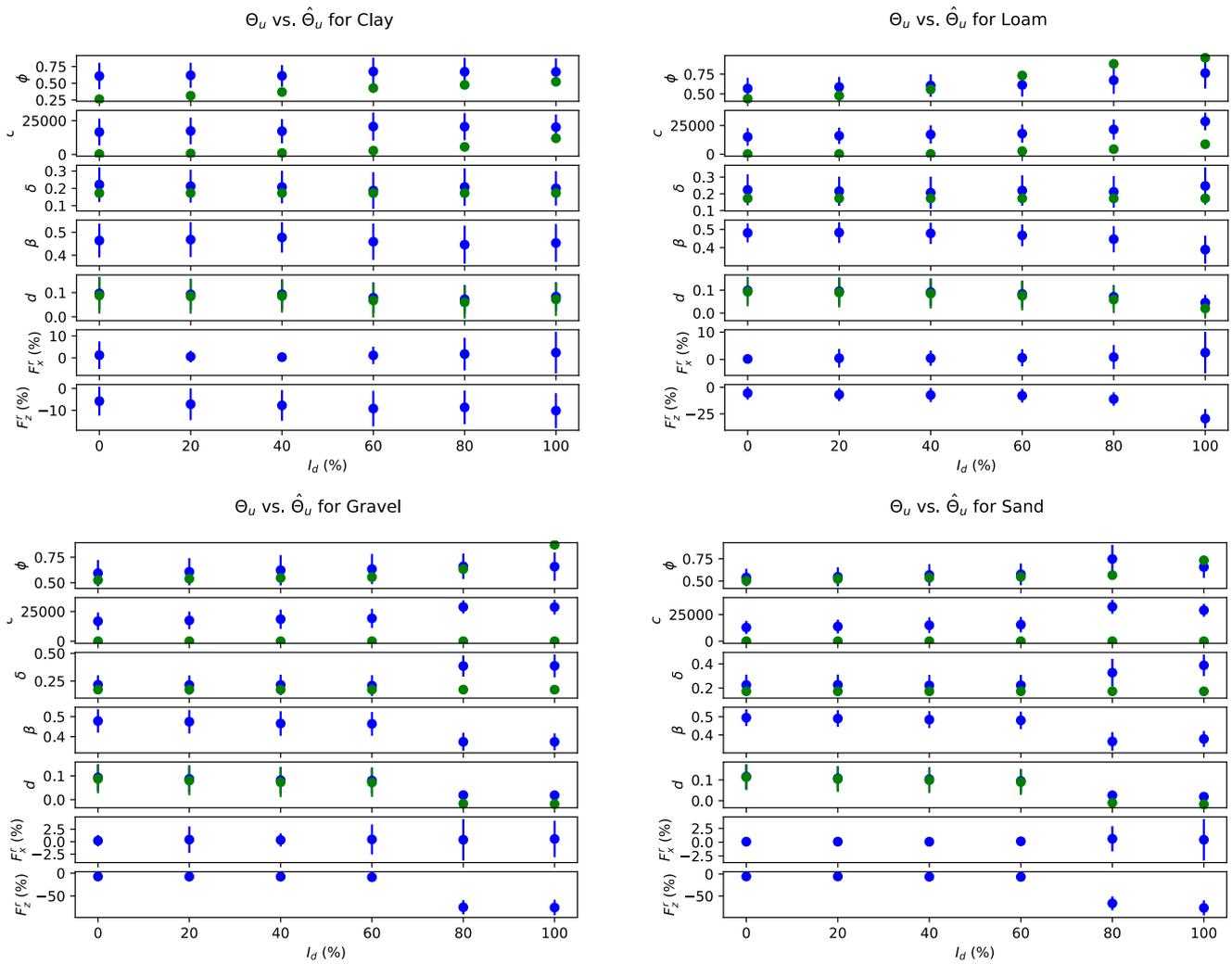

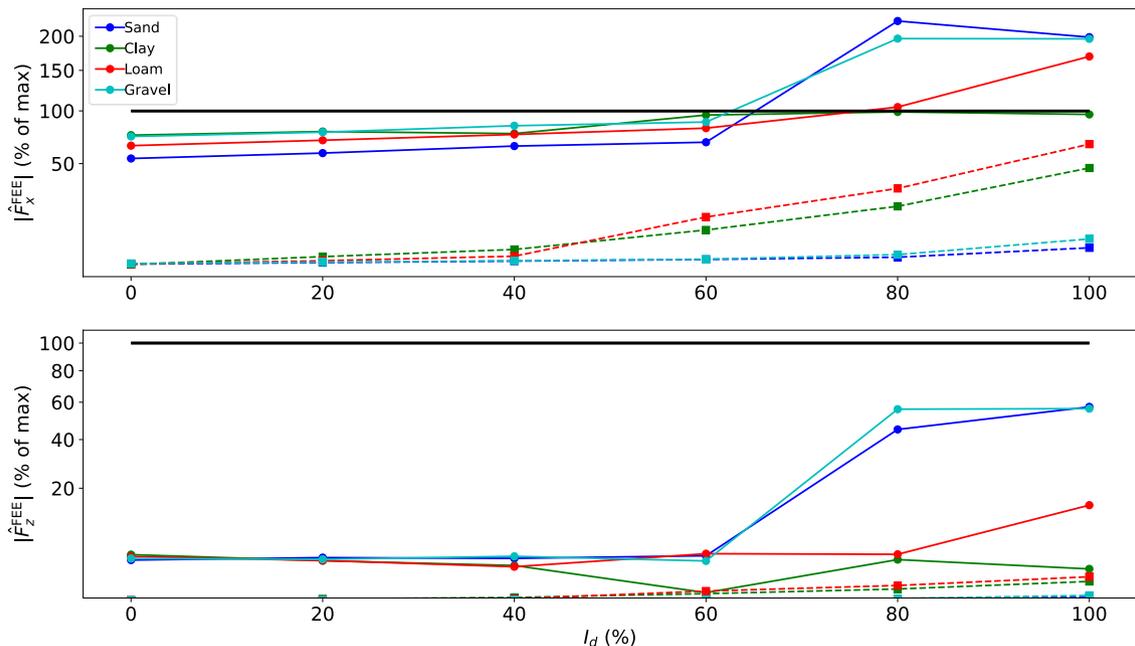

**Fig. 6** $D_{\text{default}}$ dataset (top) FEE parameters estimated by the model (blue) and the pseudo-ground-truth values obtained from Vortex. (green), (bottom) Soil failure force predicted by the FEE component of the model. Units for all angles are in radians,

For the $D_{\text{FEE}}$, we can see that the network latches on to these true parameter values. It struggles a bit in correctly estimating $\phi$, but the error is relatively small for smaller $I_d$. However, the magnitude of the residual force $F^r$ also increases correspondingly, hinting that this magnitude may be a predictor of degrading parameter estimation performance. This is intuitive because as the larger $F^r$ becomes, the less of the force is accounted for by the FEE model. Additionally, as $I_d$ increases the average depth of cut $d$ across the dataset also decreases. This is because the vehicle is experiencing higher resistance to movement and increased stalling, which forces the blade controller to raise the blade to account for poor tracking of the forward velocity command. This indicates that the assumption of soil failure made by the FEE is increasingly being violated as the soil becomes more compact. When this assumption does not hold, then the model becomes invalid, and it becomes impossible to correctly estimate parameters from the observations alone.

This analysis is useful but does not inform the effect errors in the parameters have on the estimated failure force. Small parameter error is a sufficient but not necessary condition to achieve low error of the failure force. This is because the FEE model, Eqs. 2-10, is under constrained, meaning that there are a multitude of parameter combinations that will produce the same failure force. This presents a significant challenge, particularly as the number of unknown model parameters increases. In large part, this is why the loss function, Eqs. 13-16, contains so many regularization components. These components all constrain the estimation to encourage the network to learn the true underlying parameter values.

To complement the direct parameter error analysis and provide insight regarding the effect these parameters have on the force, the mean of the parameters at each relative density are used to produce predictions of a hypothetical soil failure force for which the depth of cut is fixed, and the surcharge is zero. This means of visualizing the results provides an alternative for evaluating the quality of the parameter estimation and illustrates the effect of the ambiguity present in the FEE model (see Figs. 5 and 6).

In general, this approach enables reasoning about the effect the estimated parameters have on the expected soil failure force, which is the relevant information autonomous soil-property aware earthmoving planners require. For the $D_{\text{FEE}}$ dataset, the predicted force obtained from the averaged estimated parameters $\widehat{F}^{FEE} = f_{FEE}(\overline{\theta}|Q = 0, d = 0.1)$ agrees well with the predicted force obtained from the ground-truth parameters $F^{FEE} = f_{FEE}(\theta|Q = 0, d = 0.1)$. This indicates that while the parameter estimates do not match exactly the ground-truth, the resulting predicted force is still accurate. This is important from an autonomy perspective, because if the parameter estimates are consistently biased, meaning that the system estimates the same parameter values for the same soil conditions, and the system is able to accurately predict the cutting force, then the model can be used to map soil conditions during earthmoving and utilize this map to plan efficiently.

Moving to the real-world analog dataset $D_{\text{default}}$, the analysis becomes a bit more challenging. The parameter estimates are noisier than in the simplified dataset, and largely only $\hat{\delta}$ matches with the pseudo-ground-truth values. This is expected, due to the increased complexity of the full Vortex model that results in interaction forces only partially derived from the FEE and the pseudo-ground-truth parameters. Broadly, the forces observed in this dataset are larger than those in $D_{\text{FEE}}$, in particular $F_z$ is much larger.

That being said, for the non-cohesive soils, $\phi$ is estimated well for lower compaction. For the cohesive soils, the general trend of both $\hat{\phi}$ and $\hat{c}$ to increase with relative density is observed. This indicates that the network has uncovered a relationship between the observational history $o_{[t:t+T]}$ and these properties that can be explained by $I_d$. Essentially, this means that the model has implicitly learned to estimate $I_d$. The predicted force plots reveal a more compelling result. As the compaction increases, a gradual rise in the predicted force is observed, mirroring the smaller changes of the pseudo-ground-truth derived forces, until $I_d > 60$, where the force rises rapidly and exceed the machine tractive force limits. Note that the mean depth of cut $d$ also drops to nearly zero at $I_d = 80$, particularly for the non-cohesive soil, again indicating that soil failure assumption is not being met.

## 4. Nomenclature

| | | |
|---|---|---|
| c | Soil cohesion | [Pa] |
| $c_a$ | Soil-tool adhesion | [Pa] |
| d | Depth of cut | [m] |
| F | Cutting force | [N] |
| $N_a$ | FEE Soil-tool adhesion coeff. | [unitless] |
| $N_c$ | FEE Soil cohesion coeff. | [unitless] |
| $N_\gamma$ | FEE Soil wedge weight coeff. | [unitless] |
| $N_Q$ | FEE Surcharge coeff. | [unitless] |
| Q | Surcharge force | [N] |
| R | Soil friction and normal force | [N] |
| $\vec{x}$ | Longitudinal direction | [unit vector] |
| W | Soil wedge weight | [N] |
| w | Tool width | [m] |
| $\vec{z}$ | Vertical direction | [unit vector] |
| α | Soil surface inclination | [rad] |
| β | Soil failure angle | [rad] |
| γ | Soil moist unit weight | [N/m³] |
| δ | Soil-tool friction angle | [rad] |
| ρ | Soil-tool angle | [rad] |
| $\sigma_n$ | Normal force | [N] |
| τ | Shear force | [N] |
| φ | Soil internal friction angle | [rad] |

## 5. Conclusions

In this work, a novel physics-infused neural network approach to soil property estimation approach is introduced, a simulation is developed to enable attaining ground truth values of soil parameters, and the system is evaluated across two different datasets representing two environments with significantly different physics. The approach is shown to work well when the assumed model aligns with the environment's physics and is shown to produce meaningful parameter estimates when the model and the environmental physics deviate substantially. For both datasets, the parameter estimates enable prediction of interaction forces that are informative for the planning of autonomous earthmoving operations.

While this work provides a novel framework for soil property estimation, a number of future improvements are being considered including: estimation of parameter uncertainty, development of a soil-property mapping system, extending the model to account for changes in the surface profile by varying α, estimation of additional parameters of the FEE model (γ, Q, and $c_a$), and modifications to the network to enable supervision of the model with only knowledge of kinematic history and without interaction force measurements to enable more rapid integration on existing equipment.

## 6. Acknowledgements

The authors are grateful to Xavier Trudeau-Morin, Marek Teichmann, and Laszlo Kovacs from CM Labs for their support on the Vortex soil simulation and for Xavier's assistance in obtaining force measurements in the simulator.

## 7. References

Althoefer, K., Tan, C. P., Zweiri, Y. H., and Seneviratne, L. D., 2009. Hybrid soil parameter measurement and estimation scheme for excavation automation. IEEE Transactions on Instrumentation and Measurement, 58(10), 3633-3641.

Azulay, O. and Shapiro, A., 2021. Wheel loader scooping controller using deep reinforcement learning. IEEE Access, 9, 24145-24154.

Backman, S., Lindmark, D., Bodin, K., Servin, M., Mörk, J., and Löfgren, H., 2021. Continuous control of an underground loader using deep reinforcement learning. Machines, 9(10), 216.

Borges, P. V. K., Peynot, T., Liang, S., Arain, B., Wildie, M., Minareci, M. G., Lichman, S., Samvedi, G., Sa, I., Hudson, N., Milford, M., Moghadam, P., and Corke, P. 2022. A survey on terrain traversability analysis for autonomous ground vehicles: methods, sensors, and challenges. Field Robotics, 2(1), 1567-1627.

Bradley, D.A. and Seward, D.W., 1998. The development, control and operation of an autonomous robotic excavator. J. Intelligent and Robotic Sys.: Theory and Applications, 21, 73-97.

CM Labs Simulations, 2016. Theory guide: Vortex software's multibody dynamics engine. https://vortexstudio.atlassian.net/wiki/spaces/VSD21A/pages/2823143940/Vortex+Theory+Guide+Document (accessed 1 Mar. 2023)

Dallas, J., Jain, K., Dong, Z., Sapronov, L., Cole, M. P., Jayakumar, P., and Ersal, T., 2020. Online terrain estimation for autonomous vehicles on deformable terrains. Journal of Terramechanics, 91, 11-22.

Egli, P., Gaschen, D., Kerscher, S., Jud, D., and Hutter, M., 2022. Soil-adaptive excavation using reinforcement learning. IEEE Robotics and Automation Letters, 7(4), 9778-9785.

Fine Software, GEO5 software: table of ultimate friction factors for dissimilar materials. https://www.finesoftware.eu/help/geo5/en/table-of-ultimate-friction-factors-for-dissimilar-materials-01/. (accessed 1 Mar. 2023)

Geotechdata.info, Geotechnical Parameters, https://www.geotechdata.info/parameter, (accessed 1 Jan. 2023).

Ha, Q. P., Nguyen, Q. H., Rye, D. C., and Durrant-Whyte, H. F., 2000, Impedance control of a hydraulically actuated robotic excavator. Automation in Construction, 9, 421-435.

Haeri, A., Tremblay, D., Skonieczny, K., Holz, D., and Teichmann, M., 2020. Efficient numerical methods for accurate modelling of soil cutting operations. Proceeds of the Int. Symposium on Automation and Robotics in Construction, 37, 608-615.

Hayashi, K., Shimada, K., Ishibashi, E., Okamoto, K., and Yonezawa, Y., 2013. Development of D61EXi/PXi-23 Bulldozer with automatic control system of work equipment. Komatsu Tech. Report, 59(166).

Holz, D., Azimi, A., and Teichmann, M., 2013. Real-time simulation of mining and earthmoving operations: a level set-based model for tool-induced terrain deformations. Proceedings of the Int. Symposium on Automation and Robotics in Construction, 30.

Holz, D. 2009. Deformable Terrain for Real-Time Simulation. Thesis, RWTH Aachen University.

Iagnemma, K., Kang, S., Shibly, H., and Dubowsky S., 2004. Online terrain parameter estimation for wheeled mobile robots with application to planetary rovers. IEEE Transactions on Robotics, 20(5), 921-927.

Jackson, T., 2017. Site-by-side: comparing Cat, Deere, Komatsu, and Topcon mast-less dozer GPS systems. Equipment World, www.equipmentworld.com/side-by-side-comparing-cat-deere-komatsu-and-topcon-mast-less-dozer-gps-systems, (accessed: 1 March 2022).

Jud, D., Hottiger, G., Leemann, P., and Hutter, M., 2017. Planning and control for autonomous excavation. IEEE Robotics and Automation Letters, 2(4), 2151-2158.

Karpatne, A., Watkins, W., Read, J., and Kumar, V., 2022. Physics-guided neural networks (PGNN): an application in lake temperature modeling, in: Knowledge Guided Machine Learning, edited by Karpatne, A., Kannan, R., and Kumar, V., Chapman and Hall/CRC, New York.

Lopez-Arreguin, A. J. R., Montenegro, S., and Dilger, E. 2021. Towards in-situ characterization of regolith strength by inverse terramechanics and machine learning: A survey and applications to planetary rovers. Planetary and Space Science, 204, 105271.

Lopez-Arreguin, A. J. R. and Montenegro S., 2021. Machine learning in planetary rovers: A survey of learning versus classical estimation methods in terramechanics for in situ exploration. Journal of Terramechanics, 97, 1-17.

Luengo, O., Singh, S., and Cannon, H., 1998. Modelling and identification of soil-tool interaction in automated excavation. IEEE International Conference on Intelligent Robotic Systems, 1900-906.

Maeda, G.J. and Rye, D.C., 2012. Learning Disturbances in Autonomous Excavation. IEEE Conf. on Intelligent Robots and Autonomous Systems, 2599-2605.

MATHalino: Engineering Mathematics, Geotechnical engineering: unit weights and densities of soils. https://mathalino.com/reviewer/geotechnical-engineering/unit-weights-and-densities-soil, (accessed 1 Mar. 2023)

Mckyes, E., 1989. Lateral earth pressures, in: Developments in Agricultural Engineering, Elsevier Science Publishing Company Inc., New York, pp. 137-171.

Reece, A.R., 1964. The fundamental equation of earth-moving mechanics. Proceedings of the Institution of Mechanical Engineers, 179(6), 16-22.

Singh, S., 1995a. Learning to predict resistive forces during robotic excavation. IEEE Int. Conf. on Robotics and Automation, 2, 2102-2107.

Singh, S. 1995b. Synthesis of tactical plans for robotic excavation. Doctoral thesis, The Robotics Institute, Carnegie Mellon University.

Sotiropoulos, F. E. and Asada, H. H., 2019. A model-free extremum-seeking approach to autonomous excavator control based on output power maximization. IEEE Robotics and Automation Letters, 4(2), 1005-1012.

Tan, C.P., Zweiri, Y.H., Althoefer, K., and Seneviratne, L.D., 2005. Online soil parameter estimation scheme based on Newton-Raphson method for autonomous excavation. IEEE/ASME Tran. on Mechatronics, 10(2), 221-229.

Yamamoto, S., Nagase, H., Itogawa, H., and Kamikawa, N. 1997. Dozing system for a bulldozer. U.S. Patent No. 5950141A.

Zerveas, G., Jayaraman, S., Patel, D., Bhamidipaty, A., and Eickhoff, C., 2021. A transformer-based framework for multivariate time series representation learning. Proceedings of the ACM SIGKDD Conference on Knowledge Discovery & Data Mining, 27, 2114–2124.